\documentclass{article}
\usepackage{spconf,amsmath,graphicx}
\usepackage{multirow}
\usepackage{multicol}

\title{Enhancing Subtask Performance of Multi-modal Large Language Model}
\name{Yongqiang Zhao $^1$, Zhenyu Li $^2$, Feng Zhang $^2$, Xinhai Xu $^2$, Donghong Liu $^2$}
\address{$^1$Peking University, $^2$Academy of Military Science}

\begin{document}

%
\maketitle

\begin{abstract}
Multi-modal Large Language Model (MLLM) refers to a model expanded from a Large Language Model (LLM) that possesses the capability to handle and infer multi-modal data.
Current MLLMs typically begin by using LLMs to decompose tasks into multiple subtasks, then employing individual pre-trained models to complete specific subtasks, and ultimately utilizing LLMs to integrate the results of each subtasks to obtain the results of the task.
In real-world scenarios, when dealing with large projects, it is common practice to break down the project into smaller sub-projects, with different teams providing corresponding solutions or results. The project owner then decides which solution or result to use, ensuring the best possible outcome for each subtask and, consequently, for the entire project.
Inspired by this, this study considers selecting multiple pre-trained models to complete the same subtask. By combining the results from multiple pre-trained models, the optimal subtask result is obtained, enhancing the performance of the MLLM.
Specifically, this study first selects multiple pre-trained models focused on the same subtask based on distinct evaluation approaches, and then invokes these models in parallel to process input data and generate corresponding subtask results. Finally, the results from multiple pre-trained models for the same subtask are compared using the LLM, and the best result is chosen as the outcome for that subtask.
Extensive experiments are conducted in this study using GPT-4 annotated datasets and human-annotated datasets. 
The results of various evaluation metrics adequately demonstrate the effectiveness of the proposed approach in this paper.
\end{abstract}

\begin{keywords}
Multi-modal Large Language Model, Optimal Subtask Result, Multiple Pre-trained Models
\end{keywords}

\section{Introduction}
\label{sec:intro}
Multi-modal Large Language Models (MLLMs)~\cite{wang2023large, huang2023chatgpt, zhang2023speechgpt} have recently emerged as a new focal point of research.
They employ the potent capabilities of Large Language Models (LLMs) as cognitive engines to undertake multi-modal tasks, holding crucial significance in driving research and applications in the realm of multi-modal comprehension.
MLLMs can be applied in various domains, including medical question answering, autonomous driving, e-commerce, and more~\cite{yin2023survey, chang2023two}.
Their capacity to more effectively comprehend and process multi-modal data represents a crucial pathway towards achieving genuine artificial intelligence.

\begin{figure}[t]
\centering
\includegraphics[width=0.95\columnwidth]{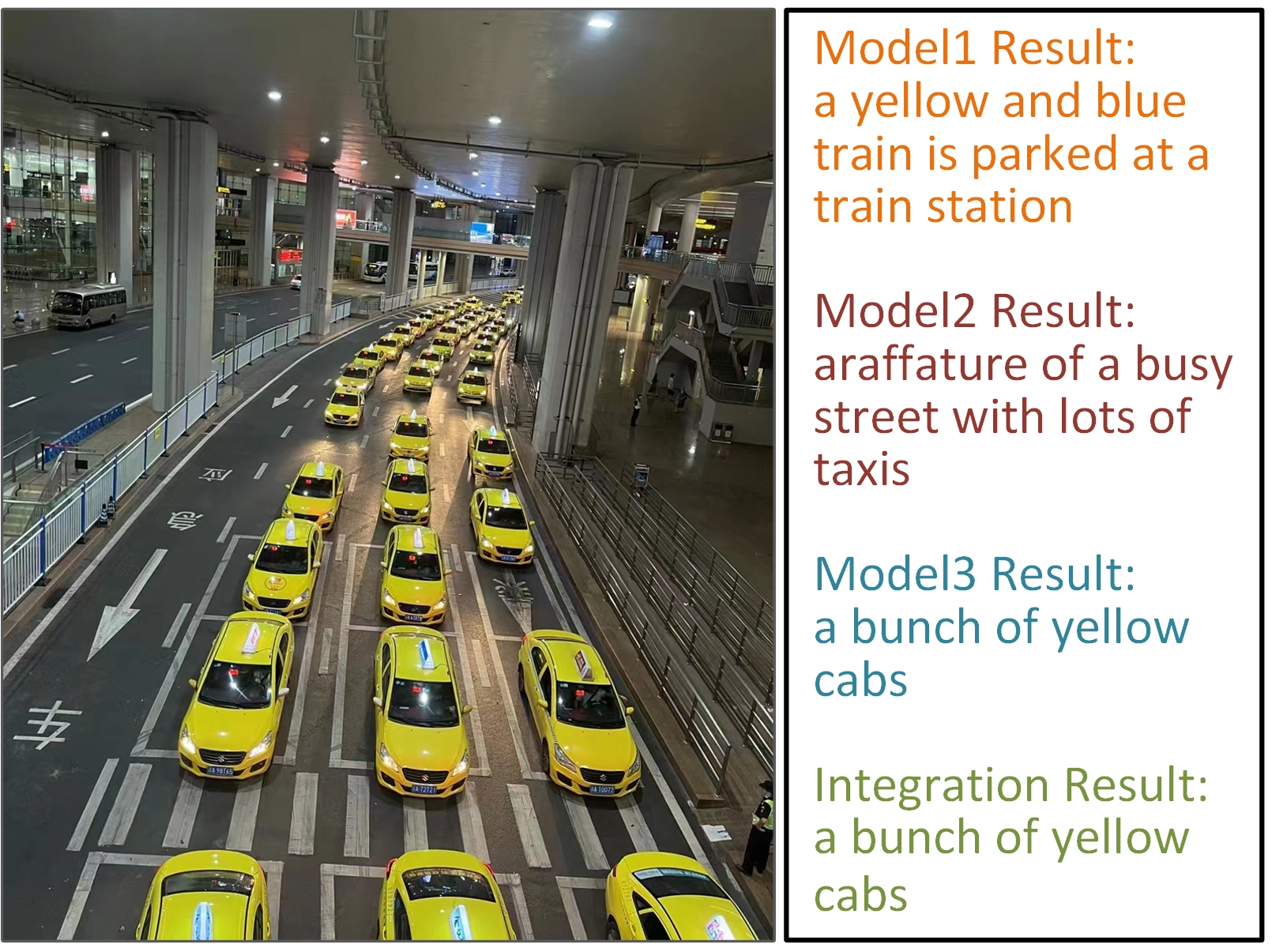}
\caption{An example of obtaining optimal result for subtask (image captioning) through the integration of outcomes from multiple pre-trained models.}
\label{fig1}
\end{figure}

Existing MLLMs~\cite{shen2023hugginggpt, wu2023visual, gupta2023visual} typically follow a process in task completion where they initially decompose tasks using LLMs, then employ a single pre-trained model to complete specific subtask, and finally, use an LLM to integrate all subtask results into the final task outcome.
However, in real-life scenarios, when a large project needs to be completed, it is rare to simply find a team that can confidently complete a subtask.
A more effective approach is to distribute the subtask and have different teams propose various solutions or results for the same subtask. Ultimately, the project's responsible entity decides which solution or outcome to adopt. This approach maximizes the potential of achieving optimal results for each subtask, ensuring the overall project's comprehensive success.
Similarly, MLLMs also struggle to achieve the best subtask result solely through the selection of a single pre-trained model.
In many cases, it is necessary to analyze the results of different pre-trained models on the same subtask to obtain the optimal subtask result.
As illustrated in Figure~\ref{fig1}, in the image captioning task, various pre-trained models generated diverse description results. Consequently, the final result selected by integrating the results of different models tends to be more accurate and reliable.

Therefore, this paper plans to utilize different pre-trained models for the same subtask during the completion of the MLLM task, resulting in diverse results. Subsequently, through analyzing these varied results, the optimal subtask result is determined. 
Specifically, this paper initiates the process by utilizing an LLM to break down the task into individual subtasks.
For a given subtask, a set of pre-trained models is acquired based on different methods. Following this, multiple pre-trained models are concurrently employed to process input data, yielding corresponding results. Subsequently, the outcomes of different pre-trained models for the same subtask are compared using the LLM, and the best result is selected. Finally, the LLM is employed to integrate all subtask outcomes, resulting in the final task result.

Our contributions can be summarized into the following two main points:
\begin{itemize}
\item A novel approach has been proposed to obtain optimal results for subtasks within MLLM. The approach involves utilizing multiple pre-trained models to accomplish the same subtask and subsequently employing an LLM to select the optimal outcome;
\item Extensive experiments are conducted on the GPT-4 annotated datasets and human-annotated datasets. The experimental results robustly demonstrate the effectiveness of the proposed method.
\end{itemize}

\section{RELATED WORK}
\label{sec:prior}
With the significant improvement in the capabilities of processing multi-modal data by models such as GPT-4~\cite{nori2023capabilities}, Kosmos-2~\cite{peng2023kosmos}, Shikra~\cite{chen2023shikra}, and HuggingGPT~\cite{shen2023hugginggpt}, MLLMs have become a focal point of research attention.
Research related to MLLMs primarily falls into four categories~\cite{yin2023survey}.
Multi-modal In-Context Learning (M-ICL): M-ICL involves utilizing a small set of examples as prompt input to stimulate the model's latent capabilities and normalize its output~\cite{dong2022survey, lu2023chameleon};
Multi-modal Chain-of-Thought (M-CoT): M-CoT obtains answers for multi-modal tasks through explicit step-by-step reasoning, presenting intermediate inference steps~\cite{wu2023visual, zhang2023multimodal};
Multi-modal Instruction Tuning (M-IT): M-IT fine-tunes a pre-trained MLLM~\cite{zhu2023minigpt, liu2023visual} by using instruction format data (instructions, multimodal inputs, and answers);
Multi-modal Tool Assistance (M-TA): M-TA involves utilizing external multi-modal tools such as search engines, OpenAI Codex, etc., to support computations beyond the core capabilities of LLMs in the reasoning process~\cite{shen2023hugginggpt, liang2023taskmatrix}.
However, existing approaches primarily yield singular results for subtasks, failing to ensure the optimality of these outcomes. Therefore, we propose a strategy that leverages multiple models to obtain subtask results and subsequently deduces the most optimal subtask outcome. This approach aims to enhance the overall performance of MLLMs.

\section{Method}
\label{sec:format}
\begin{figure*}[t]
\centering
\includegraphics[width=0.95\textwidth]{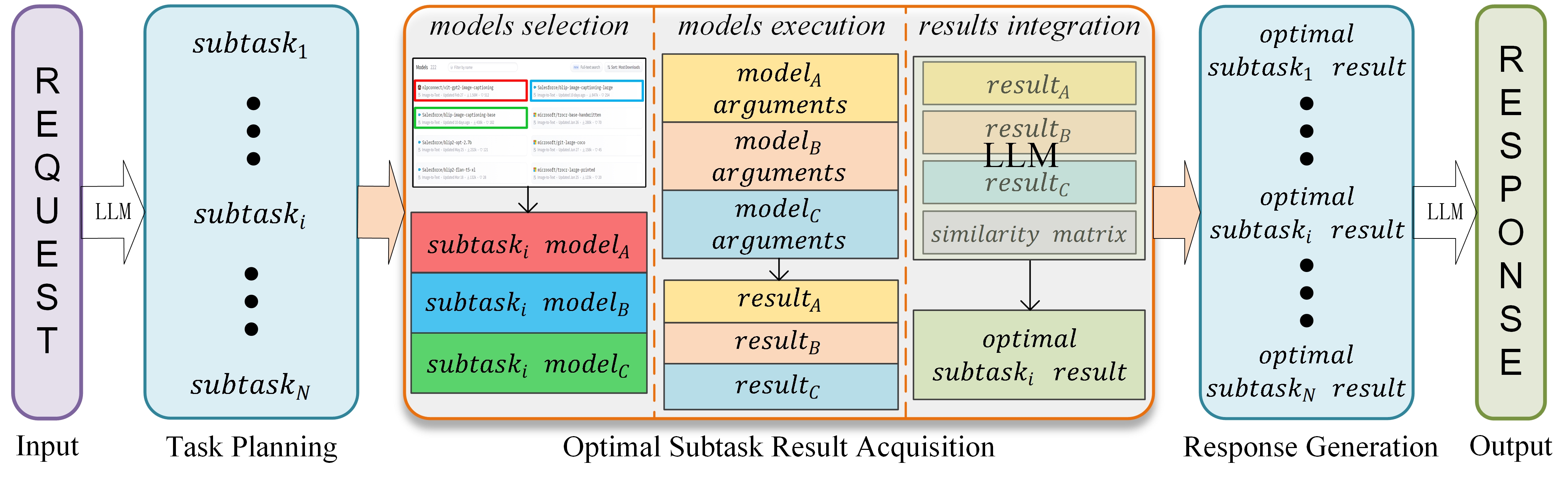}
\caption{An overview of our model architecture.}
\label{fig2}
\end{figure*}

The specific implementation process of the model ESP proposed in this paper mainly consists of three parts: task planning,  optimal subtask result acquisition, and response generation, as shown in Figure~\ref{fig2}.
Task planning: The model identifies subtask requirements from user input requests and extracts data and contextual information related to the subtasks.
Optimal subtask result acquisition: For each subtask, multiple pre-trained models are selected from a library. These models are then executed in parallel, and their results are integrated to obtain the optimal result for each subtask.
Response generation: The model summarizes and comprehensively analyzes all the returned optimal subtask results, generates the response, and returns it to the user.

\subsection{Task Planning}
In practical scenarios, user requests often involve complex intents, necessitating the coordination of multiple subtasks to achieve the overall goal.
Therefore, this paper refers to the existing methods~\cite{shen2023hugginggpt} and regards task planning as the first part of the model.
The goal is to leverage the strong commonsense understanding and logical reasoning capabilities of LLM to analyze user requests and decompose them into a set of structured subtasks.
Furthermore, this paper requires LLM to determine the dependencies and execution sequences of these decomposed subtasks to establish their interconnections.

\subsection{Optimal Subtask Result Acquisition}
\subsubsection{Models Selection}
Once task planning is complete, it is necessary to match each subtask with its corresponding model type, effectively selecting the most suitable model type for executing that subtask, such as object detection, image-to-text, image classification, etc.
To achieve this goal, this paper adopts a similar approach as HuggingGPT~\cite{shen2023hugginggpt}, utilizing model descriptions as the language interface linking model. Specifically, first, obtain the descriptions of models from the machine learning community (Hugging Face), and then dynamically select the model for each subtask through the context task-model assignment mechanism. Then the classification corresponding to the selected model is the type of model to be obtained.

After determining the model type, existing methods~\cite{wu2023visual, gupta2023visual} typically choose a single pre-trained model to execute a specific subtask. However, it is difficult for them to determine that the selected single pre-trained model is optimal, and thus cannot determine that the subtask result is optimal.
In order to ensure the best subtask result, exemplified by the machine learning community Hugging Face as an example, this paper selects multiple pre-trained models to perform the same subtask according to the rankings of multiple evaluation metrics such as "Most Downloads", "Most Likes" and "Trending". In particular, when the pre-training model is repeated, the model is selected downwards according to the corresponding evaluation metric. This approach ensures multiple high-performance pre-trained models for the same subtask.


\subsubsection{Models Execution}
After selecting multiple pre-trained models for the same subtask, the subsequent step involves executing the multiple pre-trained models.
During this stage, the system automatically inputs the parameters of the subtask into the multiple pre-trained models and executes these models to obtain the inference results.
In particular, for the same subtask, multiple pre-trained models are concurrently invoked. These multiple pre-trained models employ identical parameters, allowing for the acquisition of results from multiple pre-trained models for the same subtask.

\subsubsection{Results Integration}
After executing all the pre-trained models for the same subtask, the selection of the optimal result for the subtask becomes crucial, as it constitutes the core of the system's final decision-making.
This paper empowers LLM to receive the inference results from multiple pre-trained models for the same subtask, alongside the semantic relationships between these results. Through analyzing this information, the optimal results for the subtask are determined.

Specifically, to enhance the reliability and precision of subtask result selection, this paper employs a cosine similarity matrix~\cite{reimers2019sentence} to support LLM's decision-making process.
That is, first obtain the cosine similarity matrix between the results of each pre-trained model of the same subtask.
Subsequently, both the inference results and the cosine similarity matrix for the identical subtask are input into LLM, use LLM undertakes an analysis of all the input data, enabling the identification of the most appropriate result as the optimal result for this subtask.
It's noteworthy that the primary focus here is on selecting the most optimal individual result, rather than amalgamating existing results to generate new results. This emphasis arises due to the potential introduction of error information during the generation process.

\begin{table*}[h]
\centering
\renewcommand\arraystretch{1.45}{
\setlength{\tabcolsep}{1.68mm}{
\begin{tabular}{|c|c|c|c|c|c|c|c|c|c|c|c|c|c|}
\hline
\multicolumn{2}{|c|}{\multirow{2}*{\textbf{Model}{}{}}} & \multicolumn{4}{c|}{\textbf{Single Task}} & \multicolumn{4}{c|}{\textbf{Sequential Task}} & \multicolumn{4}{c|}{\textbf{Graph Task}}\\
\cline{3-14}
\multicolumn{2}{|c|}{} & \textbf{Acc $\uparrow$} &  \textbf{Pre $\uparrow$} & \textbf{Rec $\uparrow$} & \textbf{F1 $\uparrow$} & \textbf{ED $\downarrow$} & \textbf{Pre $\uparrow$} & \textbf{Rec $\uparrow$} & \textbf{F1 $\uparrow$} & \textbf{G4S $\uparrow$} & \textbf{Pre $\uparrow$} & \textbf{Rec $\uparrow$} & \textbf{F1 $\uparrow$}\\
\hline
\multirow{2}{*}{Alpaca-7b} & HuggingGPT & 6.48 & 35.60 & 6.64 & 4.88 & 0.83 & 22.27 & 23.35 & 22.80 & 13.14 & 16.18 & 28.33 & 20.59 \\
\cline{2-14}& ESP & \textbf{12.96} & \textbf{40.20} & \textbf{13.28} & \textbf{9.76} & \textbf{0.66} & \textbf{24.49} & \textbf{25.67} & \textbf{25.06} & \textbf{15.71} & \textbf{18.20} & \textbf{30.67} & \textbf{22.71}\\
\hline
\multirow{2}{*}{Vicuna-7b} & HuggingGPT & 23.86 & 45.51 & 26.51 & 29.44 & 0.80 & 19.15 & 28.45 & 22.89 & 19.17 & 18.66 & 13.97 & 28.08 \\
\cline{2-14}& ESP & \textbf{27.32} & \textbf{50.06} & \textbf{30.16} & \textbf{33.00} & \textbf{0.64} & \textbf{21.37} & \textbf{31.70} & \textbf{25.23} & \textbf{21.60} & \textbf{20.86} & \textbf{15.57} & \textbf{31.29}\\
\hline
\multirow{2}{*}{GPT-3.5} & HuggingGPT & 52.62 & 54.45 & 32.26 & 39.05 & 0.54 & 61.09 & 45.15 & 51.92 & 50.48 & 54.90 & 49.23 & 51.91 \\
\cline{2-14}& ESP & \textbf{54.37} & \textbf{56.23} & \textbf{33.89} & \textbf{40.02} & \textbf{0.51} & \textbf{62.98} & \textbf{46.37} & \textbf{53.47} & \textbf{52.37} & \textbf{57.01} & \textbf{51.04} & \textbf{53.88} \\
\hline
\end{tabular}}}
\caption{Experiment results of our model ESP and existing models on three tasks in the GPT-4 annotated dataset.}
\label{table1}
\end{table*}

\subsection{Response Generation}
Finally, this paper utilizes LLM to amalgamate the optimal results of all subtasks to generate a response ~\cite{shen2023hugginggpt, gupta2023visual}, which integrates all the information of the first two parts into a concise summary. 
The summary encompasses the list of planning tasks and optimal results for all subtasks.
Most important are the optimal results of the subtasks, as they constitute the pivotal points of the model's final decision.
These optimal results are presented in a structured format, such as categories with classification probabilities in an image classification model.
The model allows LLM to input these structured optimal results and generate responses in user-friendly human language.

\section{Experiments}
\label{sec:pagestyle}
\subsection{Implementation Details}
This section provides a comprehensive account of the implementation details relevant to the model proposed in this paper.
It offers an overview of the tasks and datasets used in the experiments, the evaluation metrics employed to assess model performance, and the specific configurations of the experimental parameters.

\textbf{Tasks.}
Previous work has classified existing MLLM tasks into three categories based on intent complexity: single task, sequential task, and graph task~\cite{shen2023hugginggpt}.
The single task indicates that the user's requests involve only a single subtask.
The sequential task represents that the user's requests can be decomposed into a sequence of subtasks.
The graph task signifies that the user's requests can be decomposed into multiple subtasks within a directed acyclic graph.

\textbf{Datasets.}
This paper utilizes two datasets to evaluate our model's performance across the aforementioned three tasks.
One dataset involves annotations automatically generated using GPT-4, encompassing 3,497 diverse user requests, including 1,450 single-task requests, 1,917 sequential-task requests, and 130 graph-task requests.
The other dataset is human-annotated, where expert annotators were invited to label intricate requests, totaling 46 user requests, including 24 sequential-task requests, and 22 graph-task requests.

\textbf{Evaluation Metrics.}
The evaluation metrics applied to the single task include Accuracy (Acc), Precision (Pre), Recall (Rec), and F1.
For the sequence task, the evaluation metrics consist of Edit Distance (ED)~\cite{marzal1993computation}, Precision, Recall, and F1.
For the graph task, the evaluation metrics comprise GPT-4 Score (G4S)~\cite{chiang2023vicuna}, Precision, Recall, and F1.
For all these evaluation metrics, within the corresponding tables of experimental results, an upward arrow indicates that a higher metric value signifies better model performance, while a downward arrow indicates that a smaller metric value indicates better model performance.

\textbf{Hyperparameters.}
This paper employs HuggingGPT~\cite{shen2023hugginggpt} as the baseline model and evaluates our approach against different LLMs (Alpaca-7b~\cite{taori2023alpaca}, Vicuna-7b~\cite{chiang2023vicuna}, and GPT-3.5). Additionally, the detailed prompts for the task planning and response generation stages are consistent with HuggingGPT.

\subsection{Main Results}
The experimental results on the GPT-4 annotated dataset are presented in Table~\ref{table1}.
It is evident from the table that our model demonstrates superior performance across the three tasks compared to HuggingGPT.
Specifically, in the respective tasks, our model exhibits significantly higher accuracy, precision, recall, F1 score, and GPT-4 Score than HuggingGPT, while also achieving notably lower Edit Distance values.
For instance, under the context of the Alpaca-7b LLM, our model enhances the F1 scores of the single task, sequential task, and graph task from 4.88, 22.80, and 20.59 to 9.76, 25.06, and 22.71, respectively, representing improvements of 100.0\%, 9.9\%, and 10.3\%.
In the case of the Vicuna-7b LLM, our model elevates the F1 scores of the single task, sequential task, and graph task from 29.44, 22.89, and 28.08 to 39.05, 51.92, and 51.91, respectively, demonstrating enhancements of 18.3\%, 105.5\%, and 65.9\%.
In the case of the GPT-3.5 LLM, our model raises the F1 scores for the single task, sequential task, and graph task from 39.05, 51.92, and 51.91 to 40.02, 53.47, and 53.88, respectively, demonstrating enhancements of 2.5\%, 3.0\%, and 3.8\%.

The results of the experiments on the human-annotated dataset are presented in Table~\ref{table2}.
It is evident from the table that our model outperforms HuggingGPT across the two tasks.
Specifically, for LLM Alpaca-7b, Vicuna-7b, and GPT-3.5, our model enhances the accuracy of the sequential task from 0, 7.45, and 18.18 to 2.5, 11.11, and 21.82, respectively, while decreasing the Edit Distance from 0.96, 0.89, and 0.76 to 0.72, 0.67 and 0.68, representing reductions of 25.0\%, 24.7\% and 10.5\%.
Simultaneously, our model improves the accuracy of the graph task from 4.17, 10.12, and 20.83 to 6.67, 12.82, and 23.81 under the three LLMs conditions, resulting in improvements of 60.0\%, 26.7\%, and 14.3\%, and increases its F1 score from 4.17, 7.84, and 16.45 to 6.67, 9.82, and 18.37, corresponding to enhancements of 60.0\%, 25.3\% and 11.7\%, respectively.

\begin{table}
\centering
\renewcommand\arraystretch{1.45}{
\setlength{\tabcolsep}{1.2mm}{
\begin{tabular}{|c|c|c|c|c|c|c|c|c|c|}
\hline
\multicolumn{2}{|c|}{\multirow{2}*{\textbf{Model}{}{}}} & \multicolumn{2}{c|}{\textbf{Sequential Task}} & \multicolumn{2}{c|}{\textbf{Graph Task}}\\
\cline{3-6}
\multicolumn{2}{|c|}{} & \textbf{Acc $\uparrow$} & \textbf{ED $\downarrow$} & \textbf{Acc $\uparrow$} & \textbf{F1 $\uparrow$}\\
\hline
\multirow{2}{*}{Alpaca-7b} & HuggingGPT & 0 & 0.96 & 4.17 & 4.17 \\
\cline{2-6}& ESP & \textbf{2.5} & \textbf{0.72} & \textbf{6.67} & \textbf{6.67}\\
\hline
\multirow{2}{*}{Vicuna-7b} & HuggingGPT & 7.45 & 0.89 & 10.12 & 7.84 \\
\cline{2-6}& ESP & \textbf{11.11} & \textbf{0.67} & \textbf{12.82} & \textbf{9.82} \\
\hline
\multirow{2}{*}{GPT-3.5} & HuggingGPT & 18.18 & 0.76 & 20.83 & 16.45 \\
\cline{2-6}& ESP & \textbf{21.82} & \textbf{0.68} & \textbf{23.81} & \textbf{18.37} \\
\hline
\end{tabular}}}
\caption{Experiment results of our model ESP and existing models on two tasks in the human-annotated dataset.}
\label{table2}
\end{table}

In summary, the comparison of these experimental results leads to the conclusion that our model showcases remarkable enhancements across tasks on both the GPT-4 annotated dataset and the human-annotated dataset, thereby providing compelling evidence of the effectiveness of the proposed methodology.
We also found that our method is effective when applied to various major language models mentioned above, leading to significant improvements across evaluation metrics for different tasks. Moreover, as the parameter size of the large language models decreases and their integration capacity weakens, the enhancements provided by our proposed method become even more pronounced.


\section{conclusion}
\label{sec:conclusion}
This paper considers the utilization of multiple pre-trained models to accomplish the same subtask. By integrating the results from these multiple pre-trained models, the optimal outcome for the subtask can be obtained, thereby enhancing the performance of multi-modal large language models.
Overall, the method proposed in this paper offers a novel possibility for developing complex artificial intelligence systems. It can provide more comprehensive solutions and yield superior results, thereby truly enhancing the performance of multi-modal large language models in their application.


\bibliographystyle{IEEEbib}
\bibliography{Template}

\end{document}